\documentclass{article}

\usepackage[final]{nips_2017}

\usepackage[utf8]{inputenc} % allow utf-8 input
\usepackage[T1]{fontenc}    % use 8-bit T1 fonts
\usepackage{hyperref}       % hyperlinks
\usepackage{url}            % simple URL typesetting
\usepackage{booktabs}       % professional-quality tables
\usepackage{amsfonts}       % blackboard math symbols
\usepackage{nicefrac}       % compact symbols for 1/2, etc.
\usepackage{microtype}      % microtypography

\usepackage[colorinlistoftodos]{todonotes}
\usepackage{makecell}
\usepackage{anyfontsize}
\usepackage{adjustbox}
\usepackage{caption}
\usepackage{subcaption}

\title{Natural Language Multitasking \\ 
 \large Analyzing and Improving Syntactic Saliency of Latent Representations}

\author{
  Gino Brunner, Yuyi Wang, Roger Wattenhofer, Michael Weigelt\thanks{Authors are listed in alphabetical order.} \\
  ETH Zurich, Switzerland \\
  \texttt{\{brunnegi,yuwang,wattenhofer,weigeltm\}@ethz.ch}\\
}

\begin{document}

\maketitle

\begin{abstract}

We train multi-task autoencoders on linguistic tasks and analyze the learned hidden sentence representations. The representations change significantly when translation and part-of-speech decoders are added. The more decoders a model employs, the better it clusters sentences according to their syntactic similarity, as the representation space becomes less entangled. We explore the structure of the representation space by interpolating between sentences, which yields interesting pseudo-English sentences, many of which have recognizable syntactic structure. Lastly, we point out an interesting property of our models: The difference-vector between two sentences can be added to change a third sentence with similar features in a meaningful way.

\end{abstract}

\section{Introduction}
\label{sec:intro}

Representation learning has opened the doors for many creative neural networks that learn to generate music or extract the artistic style of a painting and apply it to an arbitrary photograph (\citet{DBLP:conf/cvpr/GatysEB16}). In computational linguistics, progress has been made in neural machine translation, speech-to-text and many other applications. However, creative algorithms that write poetry, mimic an author or even develop fictional languages are sparse or non-existent. Since good representations are crucial for such creative tasks, we examine ways to improve learned representations and develop ways to measure their linguistic quality. We analyze how improvements in the syntactic capabilities of a model relate to the learned hidden representations. A syntax clustering experiment shows that the representation space of multi-task models is more easily separable into disentangled regions than that of single-task models. As a result, some simple sentence features can be added and subtracted from each other in the representation space. Our work does not focus on optimizing one single task or error, but on forcing the representations to contain useful information in a structured, analyzable way. 

To this end we train several sequence to sequence models with an increasing number of decoders. Each decoder has a distinctive linguistic task. We compare the sentence representations our models have learned and explore how representations of different sentences relate to each other.\footnote{Our code is available at \url{https://drive.google.com/open?id=0B7Mps2rt3vBoSHlWeElGT1JiS3c}}

\section{Related work}
\label{sec:rel}

\cite{DBLP:conf/nips/SutskeverVL14} use Long Short-Term Memory (LSTM) networks to encode an arbitrary sequence into a vector and decode it back into a (possibly different) sequence. They achieve results in neural machine translation (NMT) that are competitive with statistical machine translation models (SMT). The NMT objective is one of several tasks we use in our models. \citet{DBLP:journals/corr/LuongLSVK15} extend Sutskever et al.'s model with three multi-task settings: One-to-many, many-to-one and many-to-many. Their work shows that translation performance can benefit from parsing and image caption tasks. This kind of improvement from adding related tasks is also the subject of our work. \citet{DBLP:conf/wmt/NiehuesC17} note that many linguistic resources that enabled SMT are not commonly used in NMT models. They train translation models jointly with part-of-speech (POS) and named-entity recognition tasks and show that both translation and POS tagging benefit from the shared information. Our research is rooted in the same idea, but we focus on the learned representations rather than on training objectives. 

As an alternative to the bag-of-words feature many natural language processing (NLP) models use, \citet{DBLP:conf/icml/LeM14} propose the ``paragraph vector'' to represent sentences, paragraphs and whole documents. This feature outperforms bag-of-word models in text classification and sentiment analysis tasks. While it could be extended to deal with larger pieces of text, our work focuses on the sentence-level. \citet{DBLP:conf/naacl/LiuGHDDW15} developed a multi-task deep neural network for multi-domain classification and information retrieval, which learns general semantic representations useful for both tasks, demonstrating the advantage of multi-task learning. \citet{DBLP:journals/corr/abs-1710-11041} note that large parallel corpora for the training of NMT models are scarce. They introduce a novel system that solely relies on monolingual data while still learning to translate between languages. \cite{DBLP:conf/cvpr/VinyalsTBE15} develop a generative model that connects image processing and natural language generation. Their model takes an image as input and generates an English sentence that describes the content of the image. Such a complex task relies on good, well-generalized representations, which we are exploring in this work as well. 

\section{Models}
\label{sec:models}

The model architecture we use to learn representations is the autoencoder, which operates in two stages: An encoder transforms data into a ``code'' in a hidden layer, from which the decoder then tries to reconstruct the original input. The decoder can be modified to learn a task other than reconstruction, such as translating the input into a different language. Because text is of sequential nature we use Long Short Term Memory recurrent networks (LSTM) as encoders and decoders. 

\subsection{Multi-task autoencoder}

We extend the basic sequence to sequence autoencoder model by adding multiple decoders that perform separate linguistic tasks.
First, an encoder LSTM consumes a sequence of characters one by one and updates its internal state. When the whole input sequence has been read, the encoder state contains information about the entire sequence. This state is fed into a dense layer which we call the ``representation layer'', the output of which is a real vector with a specified dimension. The analysis we perform in Section~\ref{sec:res} refers to the output of this layer. Next, the representation vector is fed to each decoder LSTM, which then generate output sequences corresponding to their tasks. A single-task model will learn representations of its training data which are useful for the objective at hand. Since it supports multiple decoders, our model architecture forces the representations to contain useful information for each objective. Adding more linguistic tasks as decoders should make representations more salient from a linguistic perspective and change the properties of the whole representation space in a meaningful and analyzable way. 

\subsection{Decoders}

\begin{figure}[t]
\centering
\includegraphics[width=0.9\textwidth]{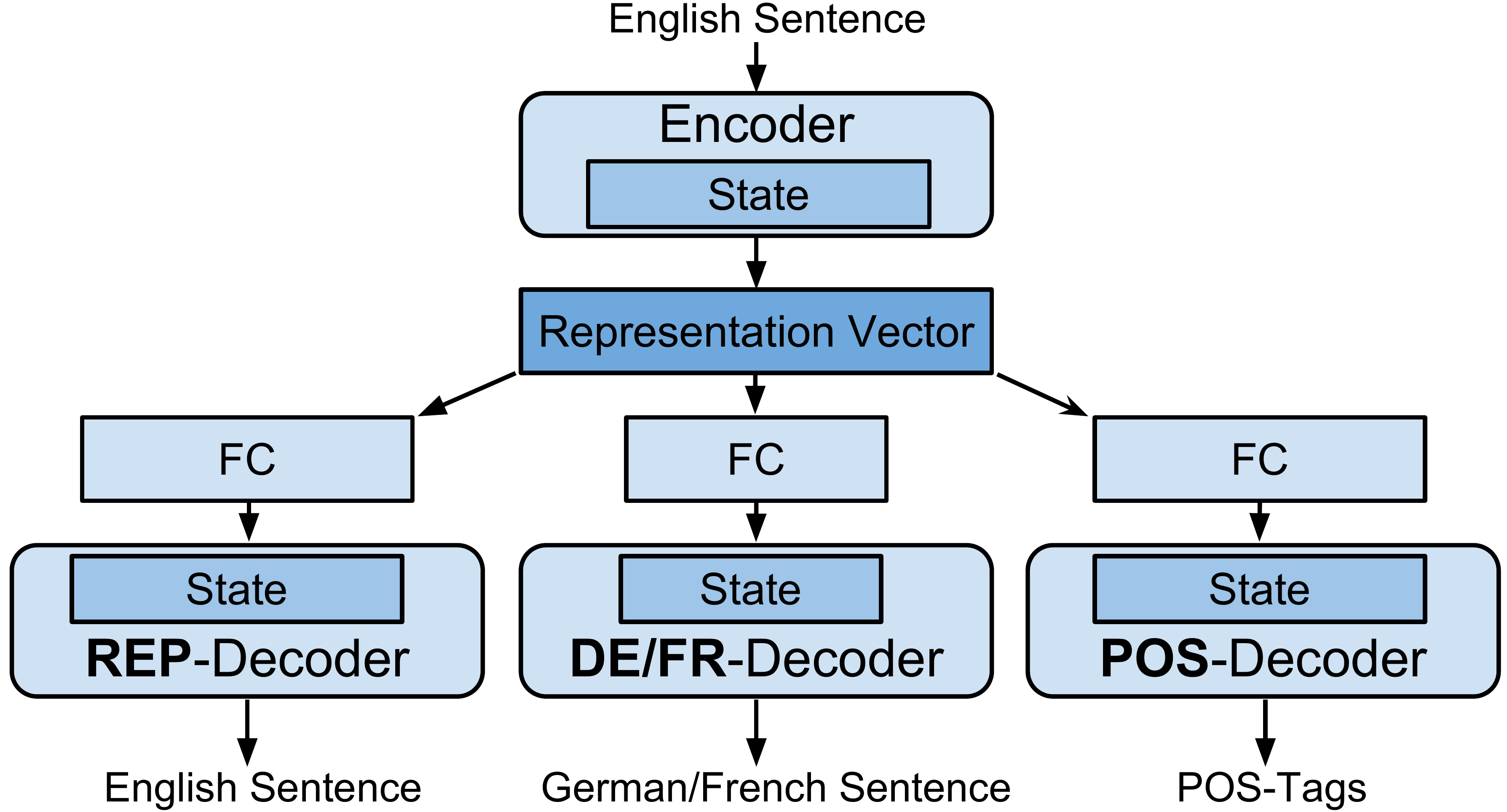}
\caption{Architecture of our multi-task autoencoder models. We use four different decoders: Replicating (REP), translation to German (DE), translation to French (FR) and a part-of-speech tagger (POS). The encoder and decoders are LSTMs. The fully connected (FC) layers transform the fixed-length representation vectors into variable length state vectors for each decoder.}
\label{fig:models}
\end{figure}

We use four different decoders.
The replicating (REP) decoder's task is to reconstruct the input sequence. The German and French (DE/FR) decoders attempt to translate the input sentence to German and French respectively. The last decoder we use learns to tag words in the input sequence with part-of-speech tags (POS), such as \emph{verb}, \emph{noun}, \emph{adjective}. Figure~\ref{fig:models} shows the architecture of our multi-task autoencoder model.

\subsection{Dataset}
\label{sub:data}
To train our models on the three tasks replication, translation and part-of-speech-tagging, we require a multilingual corpus with sentences that correspond to each other. The transcripts of the European Parliament sessions (~\citet{Europarl}) are a suitable corpus with aligned English, German and French sentences. The replication task uses the English sentences as both input and target. The training data for the POS decoders was created using the python nltk module (~\citet{nltk}) and the English text from the European Parliament corpus. The subset of this dataset we use contains over 1.7 million sentences (for all decoders), 1.5 million of which were used as the training set. The remaining 0.2 million sentences form the test set. A training example is a tuple whose size depends on the model configuration. For example, the single-task REP model uses 2-tuples (input: English sentence, target: English sentence), whereas the multi-task REP-DE-POS-model uses 4-tuples (input: English sentence, REP-target: English sentence, DE-target: German sentence, POS-target: POS-tag sequence). While the number of available training examples is the same for each model, multi-task models are trained with larger tuples and therefore more training data. To account for this imbalance, we train reference models with fewer training examples and compare their performance to the main models which we train on the full dataset. 

\subsection{Model configurations}

\label{sub:config}
We train models with different decoders and representation layer sizes. Table~\ref{tab:model} shows the perplexities reached by each decoder for some of the trained models. Perplexity measures how close a generated sequence is to a target sequence and is defined as the exponential of the cross entropy between the two sequences. The model name simply lists its decoders and representation layer size. All encoder and decoder LSTMs have 512 neurons. We have experimented with different numbers of layers, neurons and representation layer sizes. Generally, our results were not sensitive to these choices, and we thus refrained from fine-tuning our models. The achieved perplexities are not competitive with state-of-the-art models. However, our work focuses on learned representations, not on decoder losses. The perplexities reached by the reference models mentioned in~\ref{sub:data} are indistinguishable from the main models, and are thus not listed separately. 

\begin{table}[h]
  \caption{Model Configurations. For each model, the used decoderse and the size of the representation vector is indicated.}
  \label{tab:model}
  \centering
  \begin{tabular}{l l l l l}
    \toprule
Model name      & REP  & DE   & FR   & POS \\ \hline
REP-1024       & 1.05 &    - & -    & -   \\
REP-DE-1024     & 1.04 & 2.02 & -    & -   \\
REP-DE-FR-1024  & 1.04 & 2.02 & 1.86 & -   \\
REP-DE-256      & 1.07 & 2.02 & -    & -   \\
REP-DE-POS-256  & 1.14 & 2.05 & -    & 1.10\\
    \bottomrule
  \end{tabular}
\end{table}

\section{Results}
\label{sec:res}

\subsection{Syntax clustering}\label{subsec:syntaxClustering}

How can the quality of learned representations be measured?

The goal we pursue with our models is not the highest possible decoder accuracy. Instead, we are interested in representations that capture some linguistic aspect of the input language. In order to compare the learned representations of different models, we examine how well they cluster syntactically similar sentences. We use 14 sentence prototypes with different syntactic structures, for example ``The +N is +A'', where +N, +V, +A and +D are placeholders for nouns, verbs, adjectives and adverbs. Following is a list of all 13 sentence prototypes we used (the 14th category are empty sentences):

1: The +N is +A.
\newline 2:  The +N +Vs.
\newline 3: The +N has a +N.
\newline 4: The +N +Vs a +N.
\newline 5: The +N +Vs +D.
\newline 6: No +N ever +Vs.
\newline 7: Are +Ns +A?
\newline 8: The +Ns of +N +D +V the +A +N, but some +Ns still +V their +N.
\newline 9: In the +N of a +A +N, the +N will +V the +N of +Ving the +N.
\newline 10: +Ns +V the +A +N of +Ns +Ving on the +N.
\newline 11: In the +N of +N, +Ns would rather +V without +N than +V any +A +Ns.
\newline 12: +N +Vs in order to +V on a +N.
\newline 13: +A +Ns often +V like +Ns.
\newline 14: whitespace

Each sentence prototype is randomly populated by common English words 100 times. The syntax of each sentence in such a category is very similar or identical to all others in the same category, but different from sentences in other categories. These sentences are then fed into our models. We record every resulting representation and pair it with its input sentence. Using K-means clustering with $K=14$ we cluster the representation-sentence pairs in the representation space. For each resulting cluster, we count how many sentences of each prototype it contains. This yields a list such as this: $\mathtt{[30, 3, 1, 7, 88, 0, 0, 0, 0, 0, 0, 0, 0, 0]}$, which shows the content of one of 14 clusters: 30 sentences of type one, 3 of type two and so on. Since most sentences in this cluster are of type five, this cluster is assigned to be the cluster of sentence category five. However, 41 sentences of type other than five were ``falsely'' assigned to this cluster. Therefore, the error of this cluster is 41. The sum of errors of all 14 clusters is the \emph{clustering error}, which is our quality measure for this experiment. Since K-means clustering is nondeterministic, we run the algorithm 100 times. Table~\ref{tab:res} shows the best-of-100 clustering errors.

% \begin{table}[h]
%   \caption{Sentence prototypes with different syntax.}
%   \label{tab:syntaxtypes}
%   \centering
%   \begin{tabular}{l c c c c c c c}
%     \toprule
% Model & R-1024 & R-D-1024 & R-D-F-1024 & R-D-256 & R-D-P-256 \\ \hline
% % Mean  & 200 & 139 & 101 & 118 & 106 \\
% Best  & 149 & 22 & 24 & 29 & 3 \\
%     \bottomrule
%   \end{tabular}
% \end{table}

\begin{table}[h]
  \caption{Clustering errors by model}
  \label{tab:res}
  \centering
  \begin{tabular}{l c c c c c c c c}
    \toprule
% Model & R-1024 & R-D-1024 & R-D-F-1024 & R-D-256 & R-D-P-256 \\ \hline
Model & REP & REP-FR & REP-FR-DE & REP-DE & REP-POS & REP-DE-POS \\ \hline
% Mean  & 200 & 139 & 101 & 118 & 106 \\
% Best  & 149 & 22 & 24 & 29 & 3 \\
Error  & 51 & 26 & 24 & 22 & 8 & 0 \\
    \bottomrule
  \end{tabular}
\end{table}

\begin{figure}
% \centering
\begin{subfigure}{.47\textwidth}
  \centering
  \includegraphics[width=\linewidth]{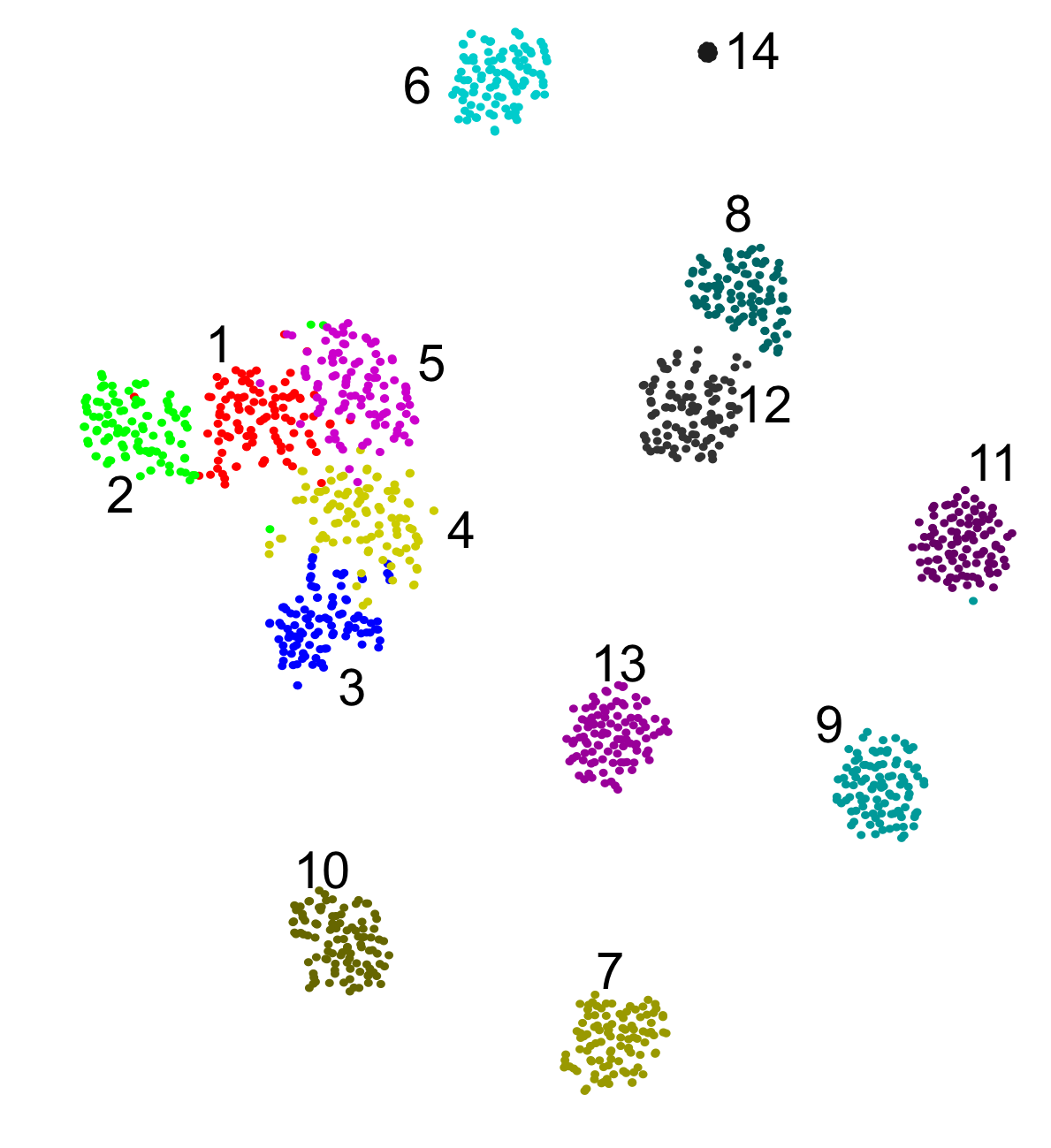}
  \caption{Model with only a replicating (REP) decoder. The sentence prototype representation clusters 1-5 as well as 8 and 12 are very close together or overlapping.}
  \label{fig:reptsne}
\end{subfigure}%
\quad \quad
\begin{subfigure}{.47\textwidth}
%   \centering
  \includegraphics[width=\linewidth]{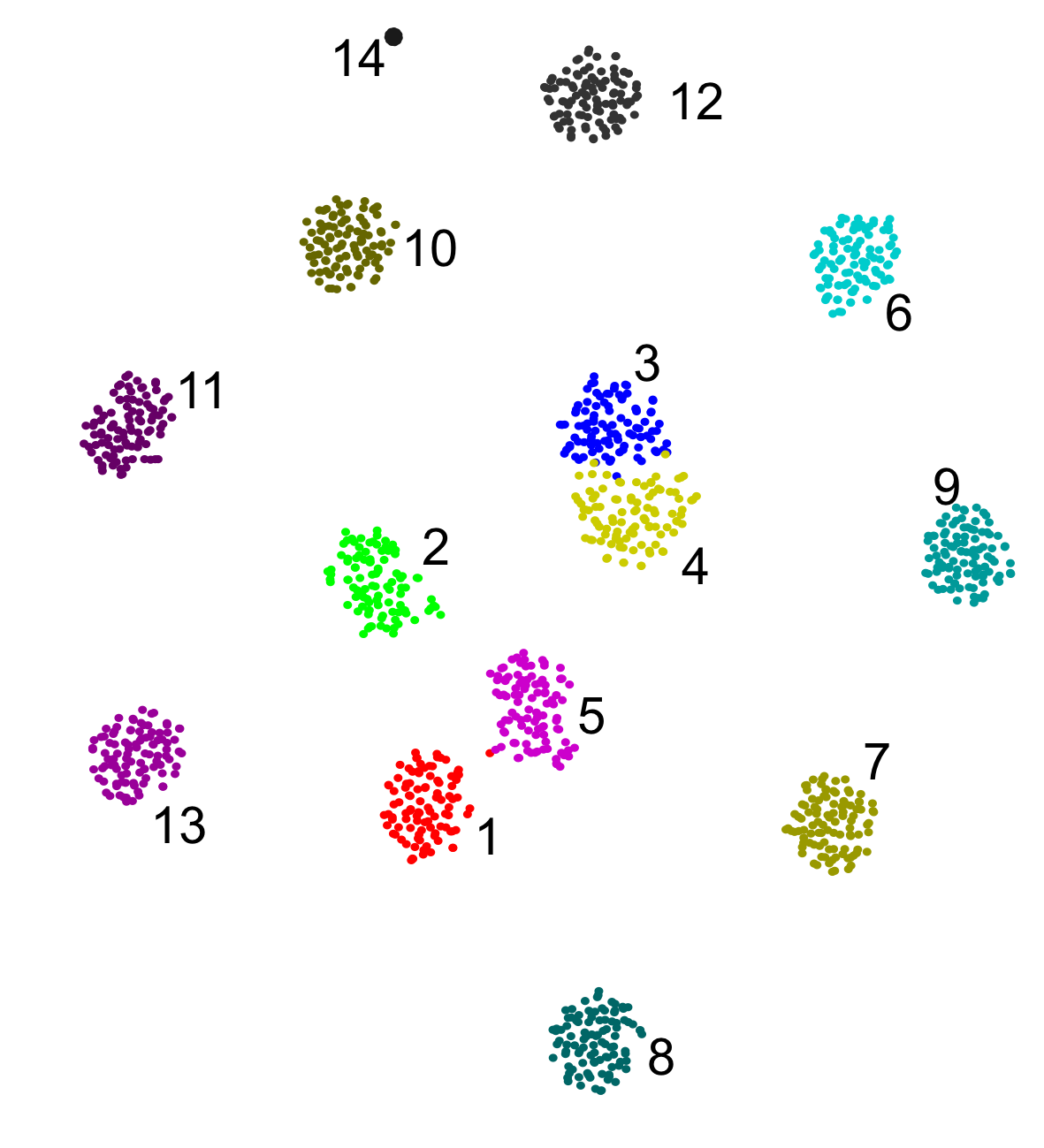}
  \caption{Model with replicating (REP), German translation (DE) and part-of-speech tagging decoders. No sentence representation clusters are overlapping, and only type 3 and 4 are close together.}
  \label{fig:repdepostsne}
\end{subfigure}
\caption{Syntax clusters visualized with t-SNE.}
\label{fig:tsne}
\end{figure}

Starting from the left, the single-task REP model has the highest clustering error. As we add German and French translation decoders, the syntax clustering error decreases. Adding part-of-speech tagging reduces the clustering error even more. Finally, combining translation and part-of-speech tagging reduces the clustering error to zero.  
% Still, the difference between a single REP and the REP-DE combination supports our hypothesis that the quality of language representations improve with additional linguistic tasks. 
% The clustering error of the rightmost model, which was trained with a POS decoder, is almost zero. 
This means that the 14 syntactically different sentence prototypes are perfectly separated in the representation space. It is not surprising that this model performs better than the others: At least for humans and most classical algorithms, correct POS tagging is a requirement or preparative step to syntax analysis. The distinctive advantage this model has over the others indicates that neural language models benefit from related linguistic tasks. Having clear, separable clusters of sentences suggests that some aspects of syntax are disentangled in the multi-task representation space. Figure~\ref{fig:tsne} shows the syntax clusters for the worst and best performing models, visualized using t-SNE. Clearly, using syntactically relevant tasks helps the model to learn more clearly separated representations of syntax. 

% \begin{figure}[t]
% \centering
% \includegraphics[width=0.9\textwidth]{NLPMT_REP_TSNE.png}
% \caption{TODO}
% \label{fig:reptsne}
% \end{figure}

% \begin{figure}[t]
% \centering
% \includegraphics[width=0.9\textwidth]{NLPMT_REP-DE-POS_TSNE.png}
% \caption{TODO}
% \label{fig:repdepostsne}
% \end{figure}

We trained several reference models on fewer training examples to account for the different numbers of decoders, and thus different amount of effective training data (as described in Section~\ref{sub:data}). Although the clustering errors differ, there is no clear trend, as some models cluster worse and others better with fewer examples (see Table~\ref{tab:refs}). Note that the perplexities these models achieved are all comparable to their reference models (trained with the full training set), and none of the models overfit the training data. 

\begin{table}[h]
  \caption{Best-of-100 clustering errors with fewer training examples}
  \label{tab:refs}
  \centering
  \begin{tabular}{l c c c c}
    \toprule
Model 				& R-D-1024 	& R-D-256 	& R-D-F-1024 	& R-D-P-256 	\\ \hline
Full training set  	& 22 		& 29 		& 24 			& 3 			\\
1/2 training set 	& 37 		& 33 		& - 			& - 			\\
1/3 training set	& - 		& - 		& 19 			& 1 			\\
    \bottomrule
  \end{tabular}
\end{table}

\subsection{Interpolation}

The sentence representations our models generate lie in a high-dimensional space. Clustering experiments show that the data points from sentences in the training set are not spaced evenly in this vector space, rather they form clusters or manifolds. Seeing how clearly some models cluster sentences according to their syntax, the question arises: What lies between two sentences? More precisely: If two sentences $s_1$ and $s_2$ have representations $r_1$ and $r_2$, what sentence corresponds to $\frac{r_1 + r_2}{2}$? What about other points along a straight line between $r_1$ and $r_2$? Table \ref{tab:ipol} shows two example outputs each of the REP decoder of the REP-DE-POS-256 and REP-DE-1024 models. As shown in Section \ref{subsec:syntaxClustering}, the REP-DE-POS-256 has a significantly lower clustering error, and it can be seen that it also produces more plausible sentences with fewer non-words when doing interpolation in the representation space. 

% We focus on the REP decoder of the REP-DE-POS-256 model, which performed best in the syntax clustering experiment. The first and last sentences in the list below are the start- and endpoints of the interpolation. The others lie equidistant on a straight line between the two.

\begin{table}[h]
  \caption{Sentence interpolation between two endpoints for two different models.}
  \label{tab:ipol}
  \begin{adjustbox}{max width = \textwidth}
  \centering
  \begin{tabular}{l l}
REP-DE-POS-256: & REP-DE-1024: \\ \hline
Is this what we want? & Is this what we want? \\
Is this meat which we need? & Is this what nath we affend? \\
Is that is very with true? & Whin to shakin the weaknes fan? \\
Whe nomists ha told day of items. & Why cust hesitage we chear thembox. \\
The course all hotels to drug itselve. & The consumer is of encautant where quote. \\
The consumer may took speed in follows Mr... & The consumer is botted there binds for EU. \\
The consumer must therefore be informed of GMOs. & The consumer must therefore be informed of GMOs \\ \hline

We will not tolerate a policy of religious repression. & We will not tolerate a policy of religious repression. \\
We will not threaten by insist regarding our representation. & We will not dossil at Ecoprat south direct-resprest and. \\
We must the knowled alro in economic objarimar represent. & We sant techni oy asplig poor correads in report'sed. \\
Whe is still want recoursion in viruccuroning responsibly. & The interest has lip porals often Mrs Martensrespre. \\
The issue where any returning ideology rrrengeing reply. & The insade to sarame places outso in requirponts resel. \\ 
There is a systemic policy of religious reprovement. & There is a strongly matic poorer 'fragen presumers' thre. \\ 
There is a systematic policy of religious repression. & There is a systematic policy of religious repression. \\
  \end{tabular}
  \end{adjustbox}
\end{table}

Unsurprisingly, not every point in the representation space corresponds to a correct English sentence. If we overlook the non-words in some of the interpolated sentences, the change in syntax can be understood to a degree. However, linear interpolation does not seem to be enough to explore the shape of this representation space. More work investigating these manifolds could yield useful results for generative, creative algorithms. For example, what properties would a representation space have if it were trained to understand two languages? How would representations of sentences and words from different languages relate to each other?

\subsection{Representation ``arithmetic''}

Some word embedding spaces have a special property, where differences between embedded words can correspond to a direction on an axis that has a semantic or grammatical interpretation. For example, the difference vector of \emph{queen} and \emph{king} might roughly equal that of \emph{woman} and \emph{man} (\citet{DBLP:conf/nips/MikolovSCCD13}). This raises the question whether sentence representations can have similar properties. A simple assumption would be that the difference-vector of \emph{Cats are good pets.} and \emph{Dogs are good pets.} should have canceled out the part about good pets and roughly point from \emph{Dogs} to \emph{Cats}. Adding this difference-vector to any sentence that contains \emph{Dogs} should then result in a sentence where \emph{Dogs} is replaced with \emph{Cats}. The REP-DE-1024 model fails at this task, as shown in Table~\ref{tab:ari_bad}. However, the arithmetic works better for the model that performed best in the syntax clustering (REP-DE-POS-256) experiment, as is shown in Table~\ref{tab:ari}. The arithmetic worked for first three sentences, and failed for the last two. While these results are not representative and require more rigorous investigation, we did not cherry pick the examples.  

The fact that the representation ``arithmetic'' works for a number of small examples shows that the learned representations are much more complex than mere symbol probabilities. Since this experiment only considers word occurrence and order, it remains open whether there is any semantic component to this phenomenon. Adding specialized semantic tasks to our models could improve the results. 

\begin{table}[h]
  \caption{Representation ``arithmetic'' for the REP-DE-1024 model.}
  \label{tab:ari_bad}
  \begin{adjustbox}{max width=\textwidth}
  \centering
  \begin{tabular}{lllllll}
$s_1$ &   & $s_2$ &   & $s_3$ &   & $s_1$ - $s_2$ + $s_3$ \\ \hline
1: I am one. & - & I am two. & + & You are two. & = & You ready no. \\
2: This example works. & - & This example fails. & + &Another attempt fails. & = &Another attempt world.\\
3: A word in a phrase. & - & A tree in a phrase. & + & A tree is green. & = &  A word is purevy? \\
4: The end is easier. & - & The start is  easier. & + & A start is next! & = & A rew lieh in new! \\
5: \makecell[l]{A large number of \\ people want to work.} & - & \makecell[l]{A small number of \\ people  want to work.} & + & \makecell[l]{A small sentence is \\ enough.} & = & \makecell[l]{A large senselfeir \\ in or evacce.}
  \end{tabular}
  \end{adjustbox}
\end{table}

\begin{table}[h]
\centering
\caption{Representation ``arithmetic'' for the REP-DE-POS-256 model.}
\begin{adjustbox}{max width=\textwidth}
  \begin{tabular}{lllllll}
%     \toprule
$s_1$ &   & $s_2$ &   & $s_3$ &   & $s_1$ - $s_2$ + $s_3$ \\ \hline
1: I am one. & - & I am two. & + & You are two. & = & You are one. \\
2: This example works. & - & This example fails. & + & Another attempt fails. & = & Another attempts work. \\
3: A word in a phrase. & -& A tree in a phrase. & + & A tree is green. & = & A word is green. \\
4: The end is easier. & - & The start is easier. & + & A start is next! & = & A need aid not! \\
5: \makecell[l]{A large number of \\ people want to work. }& - & \makecell[l]{A small number of \\ people want to work.} & + & \makecell[l]{A small sentence is \\ enough.} & = & \makecell[l]{A large sector for \\ challenge.} \\
    \hline
  \end{tabular}
     \end{adjustbox}
\label{tab:ari}
\end{table}

\section{Conclusion and Future Work}
\label{sec:conc}

We trained several multi-task autoencoders on linguistic tasks and analyzed the learned sentence representations. The representations change significantly when translation and part-of-speech tagging decoders are added. The more decoders a model uses, the better it can cluster sentence representations according to their syntactic similarity. This indicates that the space (at least the part of it that is associated with syntactic information) becomes more separable or disentangled as more tasks are added.

We explored the structure of the representation space by interpolating between sentences, which yields interesting pseudo-English sentences, many of which have recognizable syntactic structure. 
Finally, we point out an interesting property of our models' representations: The difference-vector between two sentence representations can be added to change a third sentence with similar features in a meaningful way. We call this process ``representation arithmetic'', since it allows adding and subtracting sentence features to and from other sentences. 

In the future, we want to get a better understanding of the shape of the representation space. Interpolating inside the manifold the data populates could enable creative algorithms which produce grammatical sentences by sampling from the inside of the manifold. 
Perhaps the ``representation arithmetic'' property can be made more robust by adding semantic tasks as decoders. If this behavior could be made more predictable, the representation space would have useful properties for generative models, as semantic features could be transferred between sentences. So far we have not constrained our latent space, since our focus did not lie on language generation. Nevertheless, we plan on experimenting with Variational Autoencoders due to their inherent capability to disentangle the latent space, which might enable better disentanglement of semantics and syntax. 

% \section*{References}

\small

\bibliographystyle{abbrv}
\bibliography{references}

\begin{thebibliography}{11}
\providecommand{\natexlab}[1]{#1}
\providecommand{\url}[1]{\texttt{#1}}
\expandafter\ifx\csname urlstyle\endcsname\relax
  \providecommand{\doi}[1]{doi: #1}\else
  \providecommand{\doi}{doi: \begingroup \urlstyle{rm}\Url}\fi

\bibitem[Gatys et~al.(2016)Gatys, Ecker, and Bethge]{DBLP:conf/cvpr/GatysEB16}
Leon~A. Gatys, Alexander~S. Ecker, and Matthias Bethge.
\newblock Image style transfer using convolutional neural networks.
\newblock In \emph{{IEEE} Conference on Computer Vision and Pattern
  Recognition, {CVPR} 2016}, pages 2414--2423, 2016.

\bibitem[Sutskever et~al.(2014)Sutskever, Vinyals, and
  Le]{DBLP:conf/nips/SutskeverVL14}
Ilya Sutskever, Oriol Vinyals, and Quoc~V. Le.
\newblock Sequence to sequence learning with neural networks.
\newblock In \emph{Advances in Neural Information Processing Systems, {NIPS}
  2014}, pages 3104--3112, 2014.

\bibitem[Luong et~al.(2015)Luong, Le, Sutskever, Vinyals, and
  Kaiser]{DBLP:journals/corr/LuongLSVK15}
Minh{-}Thang Luong, Quoc~V. Le, Ilya Sutskever, Oriol Vinyals, and Lukasz
  Kaiser.
\newblock Multi-task sequence to sequence learning.
\newblock \emph{CoRR}, abs/1511.06114, 2015.

\bibitem[Niehues and Cho(2017)]{DBLP:conf/wmt/NiehuesC17}
Jan Niehues and Eunah Cho.
\newblock Exploiting linguistic resources for neural machine translation using
  multi-task learning.
\newblock In \emph{Proceedings of the Second Conference on Machine Translation,
  {WMT} 2017}, pages 80--89, 2017.

\bibitem[Le and Mikolov(2014)]{DBLP:conf/icml/LeM14}
Quoc~V. Le and Tomas Mikolov.
\newblock Distributed representations of sentences and documents.
\newblock In \emph{Proceedings of the 31th International Conference on Machine
  Learning, {ICML} 2014}, pages 1188--1196, 2014.

\bibitem[Liu et~al.(2015)Liu, Gao, He, Deng, Duh, and
  Wang]{DBLP:conf/naacl/LiuGHDDW15}
Xiaodong Liu, Jianfeng Gao, Xiaodong He, Li~Deng, Kevin Duh, and Ye{-}Yi Wang.
\newblock Representation learning using multi-task deep neural networks for
  semantic classification and information retrieval.
\newblock In \emph{{NAACL} {HLT} 2015, The 2015 Conference of the North
  American Chapter of the Association for Computational Linguistics: Human
  Language Technologies}, pages 912--921, 2015.

\bibitem[Artetxe et~al.(2017)Artetxe, Labaka, Agirre, and
  Cho]{DBLP:journals/corr/abs-1710-11041}
Mikel Artetxe, Gorka Labaka, Eneko Agirre, and Kyunghyun Cho.
\newblock Unsupervised neural machine translation.
\newblock \emph{CoRR}, abs/1710.11041, 2017.

\bibitem[Vinyals et~al.(2015)Vinyals, Toshev, Bengio, and
  Erhan]{DBLP:conf/cvpr/VinyalsTBE15}
Oriol Vinyals, Alexander Toshev, Samy Bengio, and Dumitru Erhan.
\newblock Show and tell: {A} neural image caption generator.
\newblock In \emph{{IEEE} Conference on Computer Vision and Pattern
  Recognition, {CVPR} 2015, Boston, MA, USA, June 7-12, 2015}, pages
  3156--3164, 2015.

\bibitem[Koehn(2005)]{Europarl}
Philipp Koehn.
\newblock {Europarl: A Parallel Corpus for Statistical Machine Translation}.
\newblock 2005.

\bibitem[Loper and Bird(2002)]{nltk}
Edward Loper and Steven Bird.
\newblock Nltk: The natural language toolkit.
\newblock In \emph{Proceedings of the ACL-02 Workshop on Effective Tools and
  Methodologies for Teaching Natural Language Processing and Computational
  Linguistics - Volume 1}, ETMTNLP '02, pages 63--70. Association for
  Computational Linguistics, 2002.

\bibitem[Mikolov et~al.(2013)Mikolov, Sutskever, Chen, Corrado, and
  Dean]{DBLP:conf/nips/MikolovSCCD13}
Tomas Mikolov, Ilya Sutskever, Kai Chen, Gregory~S. Corrado, and Jeffrey Dean.
\newblock Distributed representations of words and phrases and their
  compositionality.
\newblock In \emph{Annual Conference on Neural Information Processing Systems,
  {NIPS} 2013}, pages 3111--3119, 2013.

\end{thebibliography}

\end{document}